\documentclass{article} 
\usepackage[preprint]{colm2026_conference}

\usepackage{microtype}
\usepackage{hyperref}
\usepackage{url}
\usepackage{booktabs}
\usepackage{amsmath}
\usepackage{graphicx}

\usepackage{lineno}

\definecolor{darkblue}{rgb}{0, 0, 0.5}
\hypersetup{colorlinks=true, citecolor=darkblue, linkcolor=darkblue, urlcolor=darkblue}

\title{Beyond Precision: Importance-Aware Recall for Factuality Evaluation in Long-Form LLM Generation}

\author{Nazanin Jafari \\
UMass Amherst\\
\And
James Allan \\
UMass Amherst\\
\And
Mohit Iyyer \\
University of Maryland \\
}

\begin{document}

\ifcolmsubmission
\linenumbers
\fi

\maketitle

\begin{abstract}
Evaluating the factuality of long-form output generated by large language models (LLMs) remains challenging, particularly when responses are open-ended and contain many fine-grained factual statements. Existing evaluation methods primarily focus on precision: they decompose a response into atomic claims and verify each claim against external knowledge sources such as Wikipedia. However, this overlooks an equally important dimension of factuality: \emph{recall} -- whether the generated response covers the relevant facts that should be included. We propose a comprehensive factuality evaluation framework that jointly measures precision and recall. Our method leverages external knowledge sources to construct reference facts and determine whether they are captured in generated text. We further introduce an importance-aware weighting scheme based on relevance and salience. 
Our analysis reveals that current LLMs perform substantially better on precision than on recall, suggesting that factual incompleteness remains a major limitation of long-form generation and that models are generally better at covering highly important facts than the full set of relevant facts.
\end{abstract}

\section{Introduction}
The factuality of LLM-generated responses remains a major challenge. LLMs can produce factually inconsistent statements with respect to the prompt or generate unsupported content (i.e., hallucinations)~\cite{min-etal-2023-factscore,wei2024long,huang2025survey,tonmoy2024comprehensive}. Traditional methods for evaluating the factuality of large language models (LLMs) predominantly rely on precision-based approaches. These methods typically involve decomposing the output of an LLM into smaller statements or discrete claims, which are then cross-verified against an external knowledge source, such as a search engine or a structured database~\cite{min-etal-2023-factscore,song-etal-2024-veriscore,wei2024long,kaggle-FACTS-leaderboard}. If a claim is supported by the external source, it is deemed factual. The overall factuality of the model is then quantified using precision, calculated as the ratio of supported claims to the total number of claims generated by the LLM. While precision-based evaluations provide valuable insights into how often an LLM produces verifiable claims, they inherently overlook an important aspect of factuality on whether the model omits critical information that should have been included. That is, these methods do not account for missed facts—instances where the LLM fails to generate relevant but factual information. Some recent works have begun to incorporate recall into factuality evaluation. \citet{wei2024long} approximates recall by dividing the number of supported claims by a target number of required facts for each input defined as the median number of claims generated by a collection of models, depend on the behavior of the evaluated models themselves. More recent approaches leverage external knowledge sources to assess factual completeness more directly. However, they either operationalize coverage at the level of query aspects or subtopics rather than fine-grained factual units \cite{samarinas2025beyond}, or evaluate comprehensiveness relative to a predefined corpus. Our work provide a comprehensive evaluation mechanism consisting of precision and recall. In this approach we introduce a fully automatic mechanism using Retrieval Augmented Generation to extract factual statements from the retrieved knowledge source given the query. We convert the retrieved passage to factual statements. Unlike previous work that treat each factual statement with the same importance, we introduce a novel importance weighted fact scoring where we score facts based on salience and relevance.


 We then conduct extensive experiments to understand when LLMs provide comprehensive outputs with respect to precision and recall in three long form generation domains of LongFact~\cite{wei2024long}, LongForm~\cite{koksal2023longform} and Biography generation dataset from FactScore benchmark~\cite{min-etal-2023-factscore}. We explore the trade-off between precision and recall based on input length. In summary our contributions include 1) Introducing a practical approach for extracting factual statements from knowledge sources using retrieval.  2) Providing a comprehensive evaluation metric consisting of factual precision and recall as well as incorporating factual importance to enhance factuality evaluation. 3) Comprehensive analysis of LLMs in different scenarios to see where they are most factual and where they fail to be factual.



\section{Related Work}

Prior work on long-form factuality evaluation has primarily focused on \emph{factual precision}, typically by decomposing generated responses into atomic claims and verifying each claim against external knowledge sources (e.g., \cite{min-etal-2023-factscore,kaggle-FACTS-leaderboard,song-etal-2024-veriscore,wei2024long,wanner-etal-2024-closer,huang2025medscore,eliav2025clatter}). While these methods provide fine-grained estimates of whether generated claims are supported, they do not directly capture whether important information is omitted from the response.

Recent work has begun to incorporate \emph{recall} or \emph{coverage} into factuality evaluation. Some approaches approximate recall using a predefined target number of facts that a response is expected to contain \citep{wei2024long}. However, such heuristics are inherently brittle, since the number of facts that should appear in a good response cannot be reliably inferred from model behavior alone; it also depends on the query, the requested level of detail, and the underlying information need. Beyond such approximations, another line of work derives coverage targets directly from the query or prompt. ICAT \citep{samarinas2025beyond} measures coverage over query aspects or subtopics, while VeriFact \citep{liu2025verifact} identifies missing facts from the prompt and generated response. Although these methods go beyond precision-only evaluation, they rely on the query or prompt to define the factual space to be covered, which is often insufficient in open-ended settings where prompts do not fully specify all relevant facts.

A more closely related line of work grounds recall evaluation in external source material. \citet{dejl2025comprehensiveness} and \citet{qi-etal-2024-long2rag} incorporate a background corpus or retrieved evidence to identify which source-grounded facts are covered by a response.  A further recent direction has begun to question the assumption that all claims should contribute equally to factuality evaluation. \citet{wanner2025all} incorporates claim relevance and importance with respect to the query into precision calculations. 

Our work therefore builds on prior research to propose a unified factuality evaluation framework that measures both precision and recall over source-grounded factual statements, while also introducing importance-aware weighting of reference facts based on relevance and salience. This allows our metric to move beyond aspect-level coverage and beyond unweighted comprehensiveness, yielding a more faithful assessment of whether an LLM response is not only factually correct, but also sufficiently complete on the facts that matter most.

\section{Methodology}

In this section, we describe our evaluation framework. We first formalize the problem, then present a pipeline for constructing a retrieval-conditioned reference fact set and extracting claims from LLM outputs. Finally, we define recall- and precision-oriented factuality metrics.

\subsection{Problem Formulation}

Given a large language model (LLM) \(M\) and an input query \(q\), let \(G_o = M(q)\) denote the long-form response generated by the model. Our goal is to evaluate the factual quality of \(G_o\) along two complementary dimensions: \emph{factual recall} and \emph{factual precision}.

Factual recall (\(F_{\mathrm{Rec}}\)) measures the extent to which \(G_o\) covers the facts that are salient and relevant to answering \(q\). Factual precision (\(F_{\mathrm{Prec}}\)) measures the extent to which the factual claims made in \(G_o\) are supported by external evidence.

We  further construct:
(i) a retrieval-conditioned reference fact set \(F = \{f_1, \dots, f_K\}\), consisting of atomic facts extracted from evidence documents retrieved for \(q\), and
(ii) a set of atomic claims \(C = \{c_1, \dots, c_L\}\), extracted from the model output \(G_o\).

The evaluation task is then to quantify (a) how well the  \(G_o\) cover the reference facts in \(F\) (recall), and (b) how well the claims in \(C\) are supported by the retrieved evidence (precision).

\subsection{Reference Fact Set Construction}

Evaluating LLM factuality requires a reliable reference set of facts relevant to the input query \(q\). We construct this set using an external knowledge source \(S\) (e.g., dataset-provided references, Wikipedia, or a search engine), from which we retrieve evidence and extract atomic facts.

Our reference fact construction pipeline consists of three stages:
(i) \emph{retrieval from the knowledge source},
(ii) \emph{atomic fact extraction}, and
(iii) \emph{fact salience scoring and reference set formation}.

\subsubsection{Retrieval from the Knowledge Source}

Given an input query \(q\), we retrieve the top-\(k\) documents from a knowledge source \(S\), yielding an evidence set
\[
E_q = \{d_1, d_2, \dots, d_k\}.
\]
These documents serve as grounding evidence for subsequent fact extraction and claim verification.


\subsubsection{Fact Extraction}

After retrieval, we construct a \emph{candidate reference fact inventory} from the evidence set \(E_q\) in a manner that promotes atomicity, uniqueness, and broad coverage of the retrieved content. We adopt the VeriScore method for fact extraction~\cite{song-etal-2024-veriscore}, in which retrieved passages are provided to an LLM with instructions to produce a list of short, independent factual statements. The extractor is explicitly prompted to (i) decompose compound statements into atomic facts, (ii) avoid redundancy, and (iii) preserve factual specificity (e.g., dates, quantities, and named entities).

To improve coverage and reduce context-window effects, fact extraction is performed at the passage/chunk level and the resulting fact lists are merged. We then apply post-processing to remove duplicate and near-duplicate facts via clustering facts into set of agglomerative clusters and extract canonical facts from each clusters. The resulting set is
\[
F (q) = \{ f_1, \dots, f_K \},
\]
which we treat as a set of atomic \emph{reference fact candidates} derived from the retrieved evidence.


\subsubsection{Fact Importance Scoring and Reference Set Formation}

Not all extracted facts are equally useful for evaluating a model's response to a query. We therefore assign each fact \(f_k \in F\) an \emph{importance score} that captures both its query-specific relevance and its broader salience with respect to the answer topic.

Concretely, for each fact \(f_k\) and query \(q\), an LLM judge assigns two discrete ratings on a 1--5 scale based on \emph{Relevance} to capture how relevant \(f_k\) is to answering the query \(q\) and \emph{Salience} to capture how central \(f_k\) is to the main topic/entity/entities implicated by \(q\), independent of any particular document's phrasing.

We normalize these scores to the unit interval,
\[
\tilde{r}_k = \frac{r_k - 1}{4}, \qquad
\tilde{s}_k = \frac{s_k - 1}{4},
\]
and define a composite importance score
\begin{equation}
    \mathrm{imp}(f_k; q)
    =
    \alpha \,\tilde{r}_k + \beta \,\tilde{s}_k,
    \label{eq:importance}
\end{equation}
where \(\alpha, \beta \ge 0\) control the trade-off between query-specific relevance and topic-level salience.

We then rank facts in \(F\) by \(\mathrm{imp}(f_k; q)\) in descending order and define a \emph{should-include} reference set
\[
F^\star(q) = \{ f_{(1)}, \dots, f_{(K^\star)} \},
\]
where \(f_{(1)}, f_{(2)}, \dots\) denote facts sorted by importance, and \(K^\star\) is either (i) a fixed budget or (ii) determined by an importance threshold. This ranked and optionally pruned set \(F^\star(q)\) serves as the reference inventory of key facts for query \(q\) and is used to compute factual recall/coverage (Section~\ref{sec:metrics}).


\subsection{Claim Extraction}

To systematically extract atomic claims from the LLM-generated output \(G_o\), similar to fact extraction, we directly adopt the claim extraction procedure used in VeriScore~\cite{song-etal-2024-veriscore}. In particular, we prompt an LLM to decompose the generated response into short, self-contained factual statements while preserving factual specificity and minimizing redundancy. The extractor is instructed to split compound statements into atomic claims whenever possible, yielding a set of atomic claims
\[
C = \{c_1, \dots, c_L\}.
\]
These extracted claims serve as the basis for assessing the factual precision (how many claims are correct). 

\subsection{Factual Precision and Recall Scoring}
\label{sec:metrics}

A factually reliable LLM should generate responses that are both (i) \emph{precise}, whether the generated claims are supported by evidence, and (ii) \emph{complete}, where the response covers the facts relevant to the query. We therefore evaluate factuality along two complementary dimensions: factual precision and factual recall.

\paragraph{Factual Precision.}
Following the verification-based evaluation paradigm used in FactScore~\cite{min-etal-2023-factscore} and VeriScore~\cite{song-etal-2024-veriscore}, we assess whether each extracted claim \(c_\ell \in C\) is supported by the retrieved evidence set \(E_q\). An LLM Judge assigns a claim verification label
\[
y^{\mathrm{prec}}_\ell \in \{\textsc{Supported}, \textsc{Contradicted}, \textsc{Not\ Supported}\},
\]
where the label indicates whether \(c_\ell\) is supported, contradicted, or not supported by the available evidence in \(E_q\).

We then define factual precision as the fraction of supported claims:
\begin{equation}
Prec_{\mathrm{F}}
=
\frac{1}{|C|}
\sum_{\ell=1}^{L}
\mathbb{1}\!\left[y^{\mathrm{prec}}_\ell = \textsc{Supported}\right].
\label{eq:fprec}
\end{equation}

In addition, we report the contradiction and not supported rates
\begin{equation}
\mathrm{C_{Rate}}
=
\frac{1}{|C|}
\sum_{\ell=1}^{L}
\mathbb{1}\!\left[y^{\mathrm{prec}}_\ell = \textsc{Contradicted}\right],
\end{equation}

\paragraph{Factual Recall.}
To assess completeness, we measure whether the generated claim set \(C\) covers the extracted facts \(F^\star(q)\). For each fact \(f_k \in F^\star(q)\), an LLM judge assigns a coverage label
\[
y^{\mathrm{cov}}_k \in \{\textsc{Covered}, \textsc{NotCovered}\},
\]
indicating whether the content of \(f_k\) is entailed or expressed by any statement in \[G_o\].

A factual recall score is then defined as
\begin{equation}
Rec_{\mathrm{F}}
=
\frac{1}{|F^\star(q)|}
\sum_{f_k \in F^\star(q)}
\mathbb{1}\!\left[y^{\mathrm{cov}}_k = \textsc{Covered}\right].
\label{eq:frec}
\end{equation}

When importance scores are available (Section~\ref{eq:importance}), we additionally compute an importance-weighted recall:
\begin{equation}
Rec_{\mathrm{F}}^{\mathrm{w}}
=
\frac{
\sum_{f_k \in F^\star(q)} \mathrm{imp}(f_k;q)\,
\mathbb{1}\!\left[y^{\mathrm{cov}}_k = \textsc{Covered}\right]
}{
\sum_{f_k \in F^\star(q)} \mathrm{imp}(f_k;q)
}.
\label{eq:weighted_frec}
\end{equation}
This weighting emphasizes coverage of high-importance facts (e.g., core biographical details) while down-weighting peripheral information.
\paragraph{Factual F1 score.}
F1 score acts as a singular scalar for combining precision and recall and is defined as harmonic mean of them. 



\section{Comprehensive LLM Factuality Evaluation}
\label{sec:evals}

In this section, we conduct a comprehensive evaluation of LLM factuality across three long-form generation domains. We first consider a setting in which all reference facts are treated as equally important. In this setting, we compare different LLMs and report results using $F1$, $Rec_F$, and $Prec_F$, introduced in section~\ref{sec:metrics}. We additionally analyze claim verification outcomes by categorizing generated claims as supported, unsupported, or contradicted. With this breakdown we provide a more informative view of model behavior on revealing how often models generate factually supported, unsupported or contradicted claims based on fact references.

We then examine factuality under a more realistic setting in which reference facts are assigned different levels of importance. This allows us to evaluate whether models preferentially capture the most important information. Finally, we investigate factuality in reasoning-oriented LLMs for one of the domains, offering a closer look at how factual accuracy and completeness interact with explicit reasoning behavior.



\subsection{Experimental Setup}
\paragraph{Datasets.}
We incorporate three benchmark domains for long-form generation that require factual knowledge: \textbf{FactScore}~\cite{min-etal-2023-factscore}, \textbf{LongFact}~\cite{wei2024long}, and \textbf{LongForm}~\cite{koksal2023longform}. \textbf{FactScore(Bio)} is a curated, human-annotated benchmark for factual accuracy assessment. It contains 183 biography prompts (entities sampled from Wikidata), with English Wikipedia used as the reference source for verification. In our setup, we use the 183 entities to construct biography-generation prompts and use Wikipedia as the primary knowledge source for retrieval and factual validation. \textbf{LongFact} consists of 2,280 prompts requiring long-form responses across 38 topics. For this benchmark, we use Google Search as the retrieval backend to collect query-conditioned evidence. To make evaluation tractable while preserving topical diversity, we randomly sample 200 prompts using topic-stratified sampling so that the topic diversity in the dataset is preserved.  \textbf{LongForm} contains approximately 24,000 (training) prompts generated from source documents using an LLM. For our experiments, we exclude prompts whose answers depend primarily on access to the original source document, and we restrict evaluation to prompts derived from Wikipedia, C4, StackExchange, and WikiHow. We then sample 200 prompts for evaluation.

\paragraph{Benchmark LLMs}
To evaluate the factuality of LLMs across different domains we selected 4 open source LLMs of \textbf{Llama-3.1-8B-Instruct}\footnote{https://huggingface.co/meta-llama/Llama-3.1-8B-Instruct},  \textbf{Llama-3.1-70B-Instruct}~\cite{dubey2024llama}\footnote{https://huggingface.co/meta-llama/Llama-3.1-70B-Instruct}, Qwen2.5-7B-Instruct~\cite{qwen2.5}\footnote{https://huggingface.co/Qwen/Qwen2.5-7B-Instruct}, Mistral-7B-instruct-v0.3\footnote{mistralai/Mistral-7B-Instruct-v0.3} and 2 closed LMs of GPT4o-mini~\cite{openai2024gpt4o} and Gemini2.5-flash-lite~\cite{comanici2025gemini}. For each dataset prompt, we prompt each model to generate a long-form response.  We then compute factual precision and recall for each response and report macro-averaged results over prompts within each domain. Details of the instructions are given in Appendix~\ref{a:factgen}.

\begin{table*}[ht]
\centering
\begin{tabular}{l ccc | ccc| ccc}
\toprule

\textbf{LLMs}& \multicolumn{3}{c}{Bio} & \multicolumn{3}{c}{LongFact}& \multicolumn{3}{c}{LongForm}\\
&  $Prec$ & $Rec$ & $F1$  &$Prec$ & $Rec$ & $F1$ &$Prec$ & $Rec$ & $F1$ \\
\cmidrule(r){1-1}

{Llama-3.1-8B-Inst} &  8.0 & 5.0& 4.0& 35.9 & 30.3 &28.2&22.7&5.3&5.3\\
  \midrule
{Llama-3.1-70B-Inst} & 14.0& 9.9  &8.3&   34.4 & 31.8 &27.4   &27.7 &8.5 &8.9 \\
  \midrule
{Qwen2.5-7B-Instruct}  & 6.3 &3.0 &3.0&  36.5& 34.7 &30.7 &   20.1&4.4 & 4.2\\
\midrule
{Mistral-7B-instruct-v0.3}  & 9.9   &4.8 &4.8&   35.5 &37.9 &32.2        &20.7&6.2&6.0 \\
\midrule
{GPT4o-mini}  & 13.3 &9.1 &8.2 &  38.6 &42.6 & 36.8     &33.4&7.2&8.6 \\
\midrule
{Gemini2.5-flash-lite}  & 14.1 &10.9 & 9.4 & 28.6 & 44.4 &30.4 &24.5&8.3&8.2 \\
\bottomrule
\end{tabular}

\caption{\small{Factuality of LLMs on three data domains. Recall is based on unweighted factual statements}}
\label{res:llm_factulaity}
\end{table*}
\subsection{Evaluation Results on LLM factuality}

\subsubsection{Factuality in LLMs (Equal Importance) }
Table~\ref{res:llm_factulaity} shows the results with respect to this experiment. In this experiment we consider each factual statement to be \emph{equally} important.From these results, we observe that factuality is strongly domain-dependent and that different models occupy distinct precision–recall operating points. On the Bio domain, Gemini-2.5-Flash-Lite achieves the best overall performance, with the highest precision and recall. On LongFact, GPT-4o-mini attains the highest F1 ($44.4\%$) however it has lowest precision with highest recall in LongFact domain, reflecting a higher-coverage generation style. On LongForm, Llama-3.1-70B-Instruct yields the highest F1, driven primarily by stronger recall relative to other models, while GPT-4o-mini achieves the highest precision but remains recall-limited.

\begin{table*}[ht]
\centering
\begin{tabular}{l ccc | ccc | ccc}
\toprule

\textbf{LLMs}& \multicolumn{3}{c}{Bio (11.4)} & \multicolumn{3}{c}{LongFact (7.4)}& \multicolumn{3}{c}{LongForm (10.2)}\\
&  $Prec$ & $Rec$  &$\rho$ &$Prec$ & $Rec$ & $\rho$ &$Prec$ & $Rec$ & $\rho$\\
\cmidrule(r){1-1}

{Llama-3.1-8B-Inst} &  8.0 & 5.0& 3.2& 35.9 & 30.3 &4.1&22.7&5.3&1.8\\
  \midrule
{Llama-3.1-70B-Inst} & 14.0& 9.9  &3.3&  34.4 & 31.8 & 4.7  &27.7 &8.5 &2.1 \\
  \midrule
{Qwen2.5-7B-Instruct}  & 6.3 &3.0 &2.8 &  36.5& 34.7 & 5.3& 20.1&4.4&2.0 \\
\midrule
{Mistral-7B-instruct-v0.3}  & 9.9   &4.8 & 2.5&  35.5 &37.9 &4.6  &20.7&6.2&1.9 \\
\midrule
{GPT4o-mini}  & 13.3 &9.1 & 2.5&  38.6 &42.6 &5.2    &33.4&7.2&1.4 \\
\midrule
{Gemini2.5-flash-lite}  & 14.1 &10.9 & 3.1& 28.6 & 44.4 &8.1& 24.5&8.3&3.1 \\
\bottomrule
\end{tabular}

\caption{\small{Tradeoff between precision and recall based on generated claims over factual statements. Average number of facts is provided in parentheses for each dataset. $\rho$ measures the ratio of number of generated claims over number of fact sets in each dataset.}}
\label{res:tradeoff}
\end{table*}

\paragraph{Claim to fact ratio and the precision--recall trade-off.}
To better interpret the precision recall results, we incorporate the claim to fact ratio in understanding their relevance to the precision and recall. $\rho = \frac{|C|}{|F|}$  is ratio of average number of claims per prompt over average number of unweighted facts. The larger the $\rho$ it suggests longer generated tokens by LLMs and more claims.  Table~\ref{res:tradeoff} illustrates the precision--recall tradeoff as a function of $\rho$. In the Bio domain, the difference between the models with the fewest and the most generated claims is approximately 10 claims. This gap increases to about 30 claims in LongFact, where Gemini-2.5-Flash-Lite produces the longest outputs, and to 15.3 claims in LongForm, again with Gemini-2.5-Flash-Lite generating the longest responses and consequently higher recall and lower precision. 

The precision--recall tradeoff is only visible for overly long generations seen in Gemini-2.5-Flash-Lite across different domains whereas this trend is not visible in other LLMs. Notably, GPT-4o-mini achieves near-maximum recall with a substantially lower $\rho$ while also attaining the highest precision, suggesting greater claim efficiency and stronger alignment with the supporting evidence. In this sense, $\rho$ helps distinguish models that improve recall mainly by generating more claims per reference fact from those that achieve a better balance through fewer but more reliable and evidence-supported claims.

Overall, longer generations are neither necessary nor sufficient for better factuality. Although greater verbosity can improve recall, it often reduces precision and yields limited gains when the main challenge is selecting and aligning the right facts.

\begin{figure*}[ht]
     \centering
    \includegraphics[width=\linewidth]{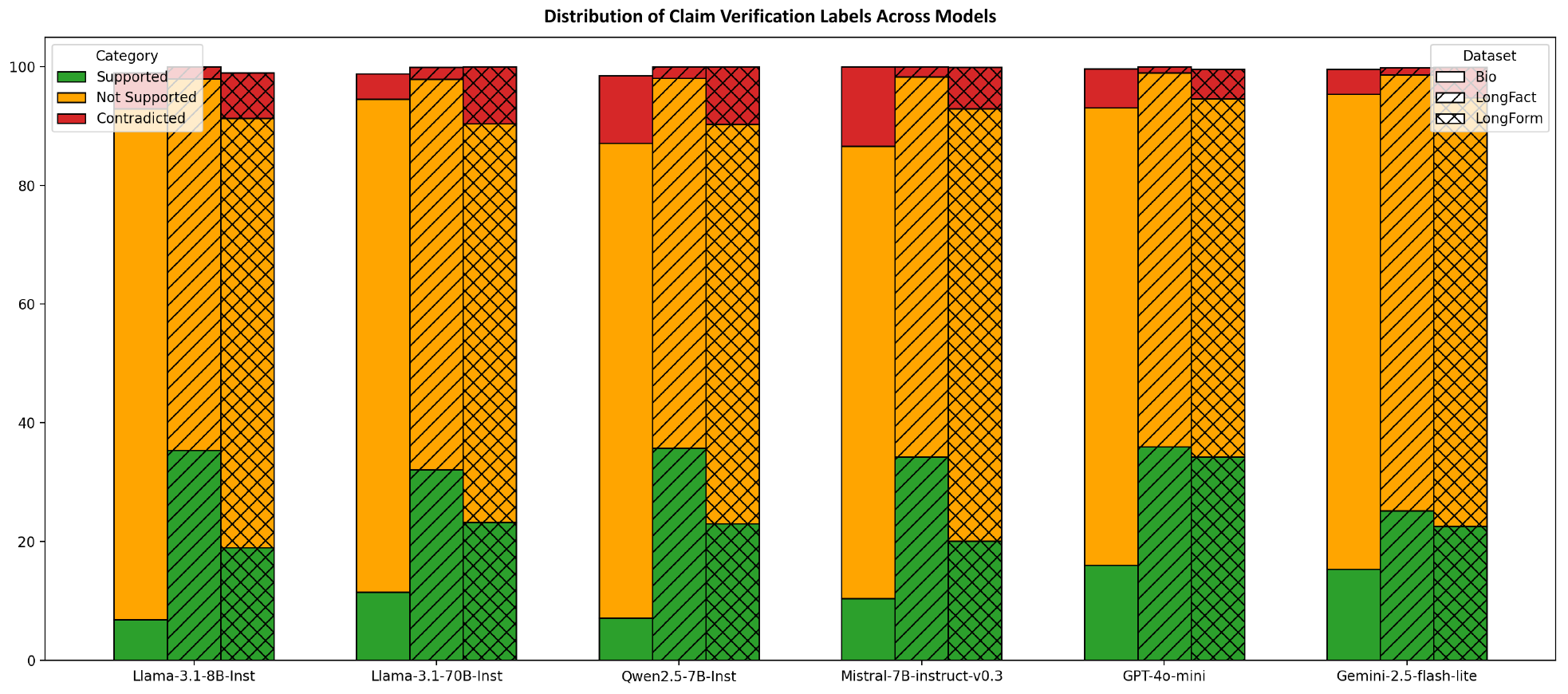}
    \caption{\small{Percentage of claims labeled as supported, not supported, or contradicted across models and datasets.}}
    \label{fig:cv_contradictions}   
\end{figure*}

\paragraph{Contradiction rates.} Beyond assessing the extent to which generated claims were supported by the retrieved evidence, we also analyzed whether any claims directly contradicted the evidence. Specifically, for each dataset, we calculated the rates of \emph{supported}, \emph{not supported}, and \emph{contradicted} claims, defined as the proportion of claims receiving each label among all generated claims in that dataset. As illustrated in Figure~\ref{fig:cv_contradictions}, \emph{not supported} claims constitute the largest category across all models and datasets, suggesting that most generated claims are not substantiated by the retrieved evidence. A smaller fraction of claims directly contradicts the evidence, with contradiction rates consistently highest on the Bio dataset across models. This is likely because errors in biography-style generation often manifest as explicit factual conflicts, such as producing an incorrect birth year when the evidence states another. 
Overall, these results suggest that factual errors in long-form generation arise more often from lack of evidential support than from direct contradiction. At the same time, the elevated contradiction rates on the Bio dataset indicate that certain domains are more prone to explicit factual conflicts.


\begin{figure*}[ht]
     \centering
    \includegraphics[width=\linewidth]{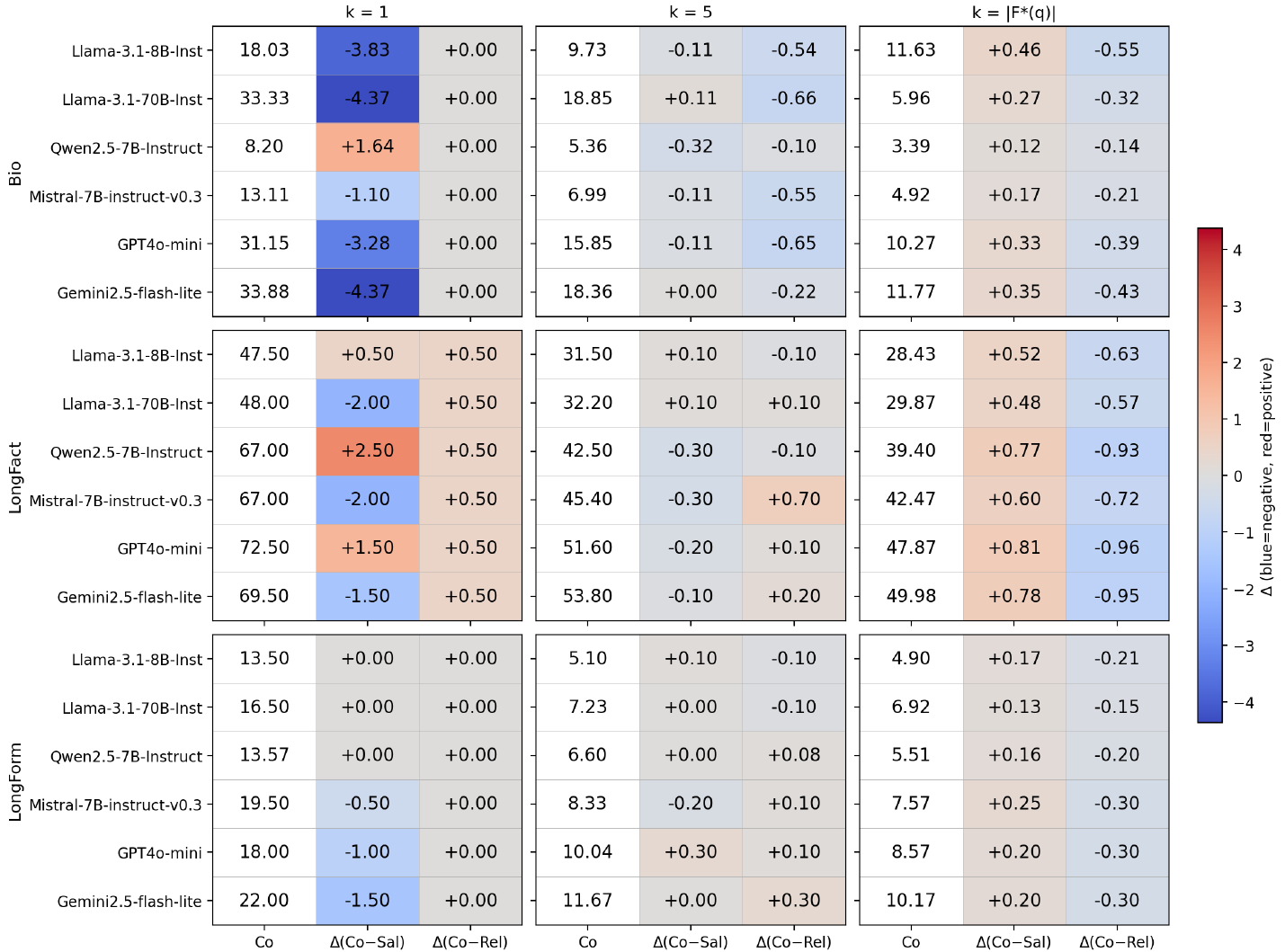}
    \caption{\small{Recall comparison for fact reference sets formed using combined importance scoring ($\alpha=\beta=1$) versus relevance-only($\alpha=1,\beta=0$) and salience-only scoring ($\alpha=0,\beta=1$). The first column in each panel reports recall for the combined score (\textbf{Co}), while the next two columns show differences relative to relevance-only (\textbf{$\Delta$(Co-Sal)}) and salience-only (\textbf{$\Delta$(Co-Rel)}) rankings.}}
    \label{fig:importance} 
\end{figure*}

\subsubsection{Factuality in LLMs (Varying Importance)} 
Figure~\ref{fig:importance} compares recall when the reference set \(F^\star(q)\) is constructed using the combined importance score (\(\alpha=\beta=1\)), relevance-only scoring (\(\alpha=1,\beta=0\)), and salience-only scoring (\(\alpha=0,\beta=1\)) across datasets and fact budgets ($K = \{1,5,|F^\star(q)|\}$. 

We observe that in all data domains, when \(K=1\), relevance-only ranking often reduces recall relative to the combined score, whereas salience-only ranking is usually comparable and sometimes slightly better, suggesting that models tend to recover the most prominent fact first. As \(K\) increases, this advantage of salience diminishes, and relevance becomes more important for broader factual coverage. At \(K=|F^\star(q)|\), the combined score consistently outperforms salience-only ranking, while relevance-only ranking often matches or slightly exceeds it, indicating that broader coverage depends more on selecting query-specific facts than on prioritizing globally salient ones. Overall, the results suggest that LLMs are relatively effective at covering a few highly important facts, but are less reliable at covering the full set of relevant facts needed for a complete answer. Overall these results support using the joint score
for reference-set formation. It preserves the strong top-fact behavior of salience, incorporates query-specific discrimination from relevance, and produces the most robust recall across datasets, models, and fact budgets.
\section{Limitations}

Despite its advantages, our framework has several limitations. First, the reliability of recall-oriented factuality evaluation depends critically on the quality of the external knowledge sources used to construct reference facts. Missing, noisy, or conflicting retrieved evidence can lead to incomplete or imperfect reference fact sets, which in turn may distort recall estimates. 

Second, our method relies on multiple automatic components, including retrieval, reference-fact extraction, claim decomposition, and fact verification. Errors in any of these stages may propagate to the final metric. 


Finally, our empirical study is limited to the models, datasets, and domains included in our experiments.

\section{Conclusion}

We presented a unified framework for factuality evaluation of long-form LLM outputs that captures both factual precision and factual recall. Our approach introduces a practical recall metric grounded in retrieval and external knowledge sources, enabling evaluation of not only whether generated claims are correct, but also whether important supported facts are missing. Empirically, we find that current LLMs generally perform much better on precision-oriented evaluation than on recall-oriented evaluation, indicating that factual incompleteness remains a key limitation of long-form generation.

In addition to overall precision and recall, our framework enables a finer-grained analysis of model errors by distinguishing supported, unsupported, and contradicted claims. Our results show that, although unsupported claims are the most common failure mode, models also produce directly contradictory statements in a meaningful subset of cases. This highlights the value of a more comprehensive evaluation framework for understanding where and how models fail.

We also introduced importance-aware factual evaluation by weighting reference facts according to relevance and salience. Our analysis shows that recall@k over the most important facts provides an informative complementary view of model quality, revealing that LLMs often capture core facts more reliably than they capture the full set of relevant facts. This further supports the need for evaluation metrics that account not only for correctness and completeness, but also for the relative importance of omitted information.

A promising direction for future work is to improve the quality of the knowledge sources used for automatic reference-fact construction, for example by filtering noisy evidence and weighting sources according to reliability. Such improvements could further strengthen the robustness and trustworthiness of automatic factuality evaluation.




\bibliography{colm2026_conference}

@misc{openai2024gpt4o,
  author    = {OpenAI},
  title     = {GPT-4o-mini Model},
  year      = {2024},
  howpublished = {\url{https://platform.openai.com/docs/models#gpt-4o-mini}},
  note      = {Accessed: January 30, 2025}
}

@article{dubey2024llama,
  title={The llama 3 herd of models},
  author={Dubey, Abhimanyu and Jauhri, Abhinav and Pandey, Abhinav and Kadian, Abhishek and Al-Dahle, Ahmad and Letman, Aiesha and Mathur, Akhil and Schelten, Alan and Yang, Amy and Fan, Angela and others},
  journal={arXiv preprint arXiv:2407.21783},
  year={2024}
}

@article{wei2024long,
  title={Long-form factuality in large language models},
  author={Wei, Jerry and Yang, Chengrun and Song, Xinying and Lu, Yifeng and Hu, Nathan and Tran, Dustin and Peng, Daiyi and Liu, Ruibo and Huang, Da and Du, Cosmo and others},
  journal={arXiv preprint arXiv:2403.18802},
  year={2024}
}

@inproceedings{song-etal-2024-veriscore,
    title = "{V}eri{S}core: Evaluating the factuality of verifiable claims in long-form text generation",
    author = "Song, Yixiao  and
      Kim, Yekyung  and
      Iyyer, Mohit",
    editor = "Al-Onaizan, Yaser  and
      Bansal, Mohit  and
      Chen, Yun-Nung",
    booktitle = "Findings of the Association for Computational Linguistics: EMNLP 2024",
    month = nov,
    year = "2024",
    address = "Miami, Florida, USA",
    publisher = "Association for Computational Linguistics",
    url = "https://aclanthology.org/2024.findings-emnlp.552/",
    doi = "10.18653/v1/2024.findings-emnlp.552",
    pages = "9447--9474"
}

@inproceedings{min-etal-2023-factscore,
    title = "{FA}ct{S}core: Fine-grained Atomic Evaluation of Factual Precision in Long Form Text Generation",
    author = "Min, Sewon  and
      Krishna, Kalpesh  and
      Lyu, Xinxi  and
      Lewis, Mike  and
      Yih, Wen-tau  and
      Koh, Pang  and
      Iyyer, Mohit  and
      Zettlemoyer, Luke  and
      Hajishirzi, Hannaneh",
    editor = "Bouamor, Houda  and
      Pino, Juan  and
      Bali, Kalika",
    booktitle = "Proceedings of the 2023 Conference on Empirical Methods in Natural Language Processing",
    month = dec,
    year = "2023",
    address = "Singapore",
    publisher = "Association for Computational Linguistics",
    url = "https://aclanthology.org/2023.emnlp-main.741",
    doi = "10.18653/v1/2023.emnlp-main.741",
    pages = "12076--12100",
}

@article{tonmoy2024comprehensive,
  title={A comprehensive survey of hallucination mitigation techniques in large language models},
  author={Tonmoy, SM and Zaman, SM and Jain, Vinija and Rani, Anku and Rawte, Vipula and Chadha, Aman and Das, Amitava},
  journal={arXiv preprint arXiv:2401.01313},
  year={2024}
}

@misc{qwen2.5,
    title = {Qwen2.5: A Party of Foundation Models},
    url = {https://qwenlm.github.io/blog/qwen2.5/},
    author = {Qwen Team},
    month = {September},
    year = {2024}
}

@article{comanici2025gemini,
  title={Gemini 2.5: Pushing the frontier with advanced reasoning, multimodality, long context, and next generation agentic capabilities},
  author={Comanici, Gheorghe and Bieber, Eric and Schaekermann, Mike and Pasupat, Ice and Sachdeva, Noveen and Dhillon, Inderjit and Blistein, Marcel and Ram, Ori and Zhang, Dan and Rosen, Evan and others},
  journal={arXiv preprint arXiv:2507.06261},
  year={2025}
}

@misc{koksal2023longform,
      title={LongForm: Effective Instruction Tuning with Reverse Instructions}, 
      author={Abdullatif Köksal and Timo Schick and Anna Korhonen and Hinrich Schütze},
      year={2023},
      eprint={2304.08460},
      archivePrefix={arXiv},
      primaryClass={cs.CL}
}

@misc{kaggle-FACTS-leaderboard,
  author = {Alon Jacovi and Andrew Wang and Chris Alberti and Connie Tao and Jon Lipovetz and Kate Olszewska and Lukas Haas and Michelle Liu and Nate Keating and Adam Bloniarz and Carl Saroufim and Corey Fry and Dror Marcus and Doron Kukliansky and Gaurav Singh Tomar and James Swirhun and Jinwei Xing and Lily Wang and Michael Aaron and Moran Ambar and Rachana Fellinger and Rui Wang and Ryan Sims and Zizhao Zhang and Sasha Goldshtein and Yossi Matias and Dipanjan Das},
  title = {FACTS Leaderboard},
  year = {2024},
  howpublished = {\url{https://kaggle.com/facts-leaderboard}},
  note = {Google DeepMind, Google Research, Google Cloud, Kaggle}
}

@inproceedings{liu2025verifact,
  title={Verifact: Enhancing long-form factuality evaluation with refined fact extraction and reference facts},
  author={Liu, Xin and Zhang, Lechen and Munir, Sheza and Gu, Yiyang and Wang, Lu},
  booktitle={Proceedings of the 2025 Conference on Empirical Methods in Natural Language Processing},
  pages={17919--17936},
  year={2025}
}

@inproceedings{qi-etal-2024-long2rag,
    title = "$LONG^{2}RAG$: Evaluating Long-Context {\&} Long-Form Retrieval-Augmented Generation with Key Point Recall",
    author = "Qi, Zehan  and
      Xu, Rongwu  and
      Guo, Zhijiang  and
      Wang, Cunxiang  and
      Zhang, Hao  and
      Xu, Wei",
    editor = "Al-Onaizan, Yaser  and
      Bansal, Mohit  and
      Chen, Yun-Nung",
    booktitle = "Findings of the Association for Computational Linguistics: EMNLP 2024",
    month = nov,
    year = "2024",
    address = "Miami, Florida, USA",
    publisher = "Association for Computational Linguistics",
    url = "https://aclanthology.org/2024.findings-emnlp.279/",
    doi = "10.18653/v1/2024.findings-emnlp.279",
    pages = "4852--4872"
}

@article{dejl2025comprehensiveness,
  title={Comprehensiveness Metrics for Automatic Evaluation of Factual Recall in Text Generation},
  author={Dejl, Adam and Barry, James and Pascale, Alessandra and Cano, Javier Carnerero},
  journal={arXiv preprint arXiv:2510.07926},
  year={2025}
}

@article{huang2025medscore,
  title={MedScore: Generalizable Factuality Evaluation of Free-Form Medical Answers by Domain-adapted Claim Decomposition and Verification},
  author={Huang, Heyuan and DeLucia, Alexandra and Tiyyala, Vijay Murari and Dredze, Mark},
  journal={arXiv preprint arXiv:2505.18452},
  year={2025}
}

@article{eliav2025clatter,
  title={Clatter: Comprehensive entailment reasoning for hallucination detection},
  author={Eliav, Ron and Cattan, Arie and Hirsch, Eran and Bassan, Shahaf and Stengel-Eskin, Elias and Bansal, Mohit and Dagan, Ido},
  journal={arXiv preprint arXiv:2506.05243},
  year={2025}
}

@article{wanner2025all,
  title={All Claims Are Equal, but Some Claims Are More Equal Than Others: Importance-Sensitive Factuality Evaluation of LLM Generations},
  author={Wanner, Miriam and Azzopardi, Leif and Thomas, Paul and Dan, Soham and Van Durme, Benjamin and Craswell, Nick},
  journal={arXiv preprint arXiv:2510.07083},
  year={2025}
}

@inproceedings{wanner-etal-2024-closer,
    title = "A Closer Look at Claim Decomposition",
    author = "Wanner, Miriam  and
      Ebner, Seth  and
      Jiang, Zhengping  and
      Dredze, Mark  and
      Van Durme, Benjamin",
    editor = "Bollegala, Danushka  and
      Shwartz, Vered",
    booktitle = "Proceedings of the 13th Joint Conference on Lexical and Computational Semantics (*SEM 2024)",
    month = jun,
    year = "2024",
    address = "Mexico City, Mexico",
    publisher = "Association for Computational Linguistics",
    url = "https://aclanthology.org/2024.starsem-1.13/",
    doi = "10.18653/v1/2024.starsem-1.13",
    pages = "153--175"
}

@inproceedings{samarinas2025beyond,
  title={Beyond factual accuracy: Evaluating coverage of diverse factual information in long-form text generation},
  author={Samarinas, Chris and Krubner, Alexander and Salemi, Alireza and Kim, Youngwoo and Zamani, Hamed},
  booktitle={Findings of the Association for Computational Linguistics: ACL 2025},
  pages={13468--13482},
  year={2025}
}

@article{huang2025survey,
  title={A survey on hallucination in large language models: Principles, taxonomy, challenges, and open questions},
  author={Huang, Lei and Yu, Weijiang and Ma, Weitao and Zhong, Weihong and Feng, Zhangyin and Wang, Haotian and Chen, Qianglong and Peng, Weihua and Feng, Xiaocheng and Qin, Bing and others},
  journal={ACM Transactions on Information Systems},
  volume={43},
  number={2},
  pages={1--55},
  year={2025},
  publisher={ACM New York, NY}
}
\bibliographystyle{colm2026_conference}

\appendix
\section{Fact Generation, Coverage and Scoring Prompts}\label{a:factgen}
Prompt instructions for generating facts form retrieved documents, evaluating coverage and scoring atomic facts: 
\begin{table*}[h]
\centering
\caption{Prompt used for fact generation.}
\label{a:factsgen_prompt}
\small
\renewcommand{\arraystretch}{1.1}
\setlength{\tabcolsep}{6pt}

\begin{tabular}{@{}p{0.96\textwidth}@{}}
\toprule
\textbf{Instructions for Fact generation}

\textit{You need to extract as many facts from the given ``context'' as possible. Output each fact in bullet-point format. Each of these facts should be generated directly from the ``context'', should be objective and factual, so avoid opinion-based sentences. Each fact should add new information and avoid redundancy. Each fact should be self-contained and make sense on its own. Choose facts that provide new insights or something unusual about the topic. Avoid general statements that apply to many things. A good fact should be specific and unique and contribute to the essential understanding of the topic.}

\medskip
\textit{Here is an example:}

\medskip
\textit{Context: Adam Jared Brody (born December 15, 1979) is an American actor. His breakout role was as Seth Cohen on the Fox television series \textit{The O.C.} (2003--2007). For his performance as Noah in the Netflix romantic comedy series \textit{Nobody Wants This} (2024), he earned a nomination for the Golden Globe Award for Best Actor in a Television Series (Musical/Comedy) and won the Critics' Choice Television Award for Best Actor in a Comedy Series.}

\medskip
\textit{Output:}

\textit{Facts:}
\textit{
\begin{itemize}
    \item Adam Jared Brody was born on December 15, 1979.
    \item Adam Brody's breakout role was as Seth Cohen on the Fox television series \textit{The O.C.} (2003--2007).
    \item He earned a nomination for the Golden Globe Award for Best Actor in a Television Series (Musical/Comedy) for his performance as Noah in the Netflix romantic comedy series \textit{Nobody Wants This} (2024).
    \item Brody won the Critics' Choice Television Award for Best Actor in a Comedy Series for his role in \textit{Nobody Wants This}.
\end{itemize}
}
\\
\bottomrule
\end{tabular}

\end{table*}

\begin{table*}[ht]
\centering
\caption{Prompt used for fact coverage verification.}
\label{tab:fact_coverage_prompt}
\small
\renewcommand{\arraystretch}{1.1}
\setlength{\tabcolsep}{6pt}

\begin{tabular}{@{}p{0.96\textwidth}@{}}
\toprule
\textbf{Instructions for fact coverage verification}

\textit{You are checking whether a given fact is COVERED by a set of claim sentences (an answer).}

\medskip
\textit{Important:}
\begin{itemize}
    \item The fact is assumed to be TRUE. Do not question or evaluate its truth.
    \item Decide whether the fact is stated or clearly implied by the claim sentences.
\end{itemize}

\medskip
\textit{Definitions:}
\begin{itemize}
    \item The fact is \textbf{COVERED} if the claim sentences clearly state or entail the fact. That means: if a careful reader had \textbf{only} these claim sentences and nothing else, they would be confident that the fact is true.
    \item The fact is \textbf{NOT\_COVERED} if the claim sentences do not provide enough information to guarantee the fact. It is \textbf{NOT\_COVERED} even if the fact seems plausible based on outside knowledge.
\end{itemize}

\medskip
\textit{Notes:}
\begin{itemize}
    \item You may use \textbf{multiple} claim sentences together to decide if the fact is covered.
    \item Do \textbf{not} use any outside knowledge beyond the claim sentences.
    \item Be strict: if the claim sentences are compatible with the fact but do not actually say or entail it, label it \textbf{NOT\_COVERED}.
\end{itemize}

\medskip
\textit{Fact:}

\texttt{\{fact\}}

\medskip
\textit{Claim sentences (numbered, 1-based indices):}

\texttt{\{claims\_block\}}

\medskip
\textit{Your tasks:}
\begin{itemize}
    \item Decide whether the fact is \textbf{COVERED} or \textbf{NOT\_COVERED} by the claim sentences above.
    \item If and only if the fact is \textbf{COVERED}, list all claim IDs (1-based indices) that directly help cover/entail the fact. If \textbf{NOT\_COVERED}, use an empty list.
\end{itemize}

\medskip
\textit{Output \textbf{strictly} as JSON, with no extra text:}

\begin{verbatim}
{
  "label": "COVERED" | "NOT_COVERED",
  "evidence_claim_ids": [<int>, ...]
}
\end{verbatim}
\\
\bottomrule
\end{tabular}

\end{table*}

\begin{table*}[ht]
\centering
\caption{Prompt used for relevance and salience scoring.}
\label{tab:relevance_salience_prompt}
\small
\renewcommand{\arraystretch}{1.1}
\setlength{\tabcolsep}{6pt}

\begin{tabular}{@{}p{0.96\textwidth}@{}}
\toprule
\textbf{Instructions for relevance and salience scoring}

\medskip
\textit{You are scoring each statement based on its relevance and salience to the given query.}

\medskip
\textit{Task:} \\
\textit{Given a query and a list of sentences about that query, assign each sentence:} \\
\textit{- a RELEVANCE rating from 1 to 5} \\
\textit{- a SALIENCE rating from 1 to 5}

\medskip
\textit{Definitions:} \\
\textit{- Relevance (1--5): how directly this sentence helps answer the query.} \\
\textit{\hspace*{1em}1 = completely unrelated} \\
\textit{\hspace*{1em}2 = weakly related} \\
\textit{\hspace*{1em}3 = somewhat related} \\
\textit{\hspace*{1em}4 = strongly related} \\
\textit{\hspace*{1em}5 = directly answers the query or is crucial to the answer}

\medskip
\textit{- Salience (1--5): how important this sentence is for answering the query among all the sentences provided.} \\
\textit{\hspace*{1em}1 = trivial detail, almost never needed} \\
\textit{\hspace*{1em}2 = minor detail} \\
\textit{\hspace*{1em}3 = useful but not central} \\
\textit{\hspace*{1em}4 = important detail that should usually be included} \\
\textit{\hspace*{1em}5 = essential; leaving it out would seriously harm the answer}

\medskip
\textit{Query:} \\
\texttt{\{query\}}

\medskip
\textit{Sentences:} \\
\texttt{\{sentence\_list\}}

\medskip
\textit{Output strictly as a list of JSON objects with this schema, and do not include any text before or after the JSON.}

\medskip
\textit{Output:} \\
\texttt{[ \{} \\
\texttt{\hspace*{1em}"id": <sentence\_index>,} \\
\texttt{\hspace*{1em}"sentence": "<sentence text>",} \\
\texttt{\hspace*{1em}"relevance": <int 1-5>,} \\
\texttt{\hspace*{1em}"salience": <int 1-5>} \\
\texttt{\}, ... ]}

\medskip




\\
\bottomrule
\end{tabular}
\end{table*}

\end{document}